\PassOptionsToPackage{table}{xcolor}
\documentclass{article} % For LaTeX2e
\usepackage[T1]{fontenc}
\usepackage{newunicodechar}
\newunicodechar{–}{\textendash}
\newunicodechar{“}{\textquotedblleft}
\newunicodechar{”}{\textquotedblright}
\usepackage{iclr2027_conference,times}

\usepackage{hyperref}
\usepackage{url}
\usepackage{graphicx}
\usepackage{amsmath}
\usepackage{amssymb}
\usepackage{array}
\usepackage{tabularx}
\usepackage{multirow}
\usepackage[table]{xcolor}  
\usepackage{booktabs}
\usepackage{makecell}
\usepackage{caption}
\usepackage{subcaption}    
\usepackage{algorithm}
\usepackage{algpseudocode}
\usepackage{soul}         
\usepackage{wrapfig}

\definecolor{citeblue}{RGB}{0,114,178}
\hypersetup{
    colorlinks=true,
    citecolor=citeblue,
    linkcolor=blue
}

\title{
\textcolor{orange!85!black}{Tac}\textcolor{gray!70!black}{2}\textcolor{blue!70!black}{Pix}:
Image-Space Visuo-Tactile Fusion for Dexterous Manipulation
}

\author{
\textbf{Shengcheng Luo}$^{1,2*}$,
\textbf{Kefei Wu}$^{1*}$,
\textbf{Xiaoying Zhou}$^{1}$, 
\textbf{Wanlin Li}$^{2}$, 
\textbf{Ziyuan Jiao}$^{2\dagger}$,
\textbf{Chenxi Xiao}$^{1\dagger}$ \\[0.5em]
$^{1}$ShanghaiTech University, 
$^{2}$Beijing Institute for General Artificial Intelligence 
}

\makeatletter
\newcommand{\authornote}[1]{%
  \begingroup
  \def\@thefnmark{}%
  \def\@makefntext##1{\noindent##1}%
  \@footnotetext{#1}%
  \endgroup
}
\makeatother
\iclrfinalcopy % Uncomment for camera-ready version, but NOT for submission.
\begin{document}
\authornote{%
$^{*}$Equal contribution.\qquad
$^{\dagger}$Corresponding authors.%
}

\maketitle

\vspace{-20bp}

\begin{center}
\includegraphics[width=\textwidth]{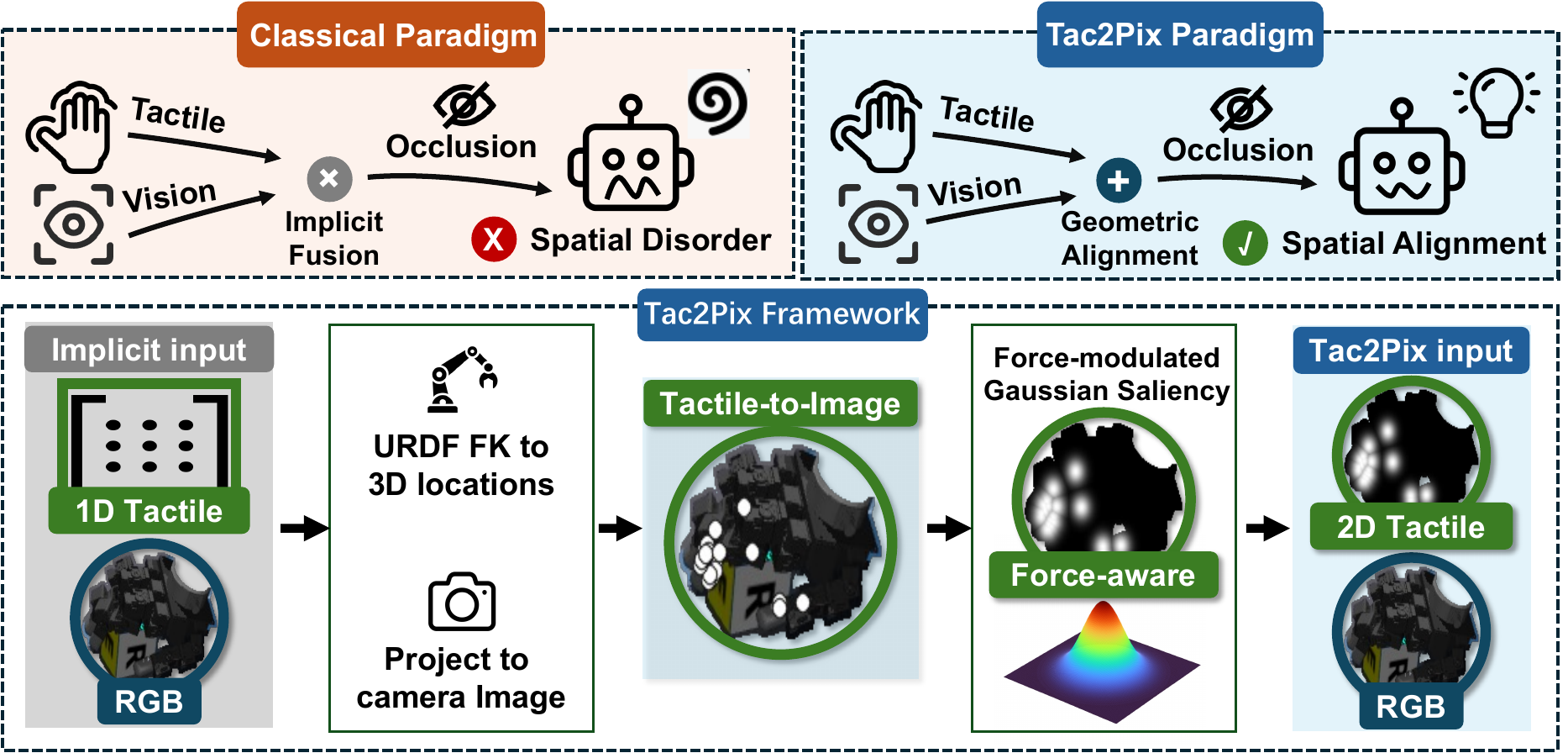}
\captionof{figure}{\textbf{Overview of Tac2Pix}. Classical tactile-vision fusion relies on implicit multimodal embeddings that often lose spatial correspondence under occlusion. Our \textbf{Tac2Pix} paradigm explicitly projects tactile contacts onto image-space saliency maps, producing a force-aware and spatially aligned representation for robust dexterous manipulation.}
\label{fig:teaser}
\end{center}

\begin{abstract}
Vision and touch are naturally complementary for dexterous manipulation. Yet they are represented in fundamentally different forms: visual observations are dense and spatially structured, whereas tactile measurements are sparse, heterogeneous, and robot-centric. This mismatch makes it challenging to establish explicit spatial correspondence between the two modalities, particularly when they are fused in latent space. However, each tactile measurement is associated with a known location on the robot’s kinematic chain, offering a localization prior that could bridge this representational gap. This raises a natural question: \emph{can this localization prior be used to construct a shared representation that explicitly aligns touch with vision?}
We introduce \textbf{Tac2Pix}, an image-space visuo-tactile fusion framework that projects sparse tactile measurements onto the RGB image plane using forward kinematics and camera calibration, then renders them as force-aware saliency maps. By expressing touch in the same pixel space as vision, Tac2Pix directly leverages pretrained 2D visual representations. We further employ zero-initialized input adaptation to preserve the original visual pathway at the start of policy learning.
To assess whether this geometry-based interface yields consistent policy gains, we evaluate Tac2Pix on three simulated and three real-world dexterous manipulation tasks. In simulation, Tac2Pix consistently improves visuo-tactile policy learning across tasks, policy architectures, and visual backbones. In real-world experiments, it improves average success under visual occlusion by $26.7$ percentage points over the strongest latent visuo-tactile fusion baseline, with gains persisting under both random and physical occlusions. Together, these results support image space as an effective interface for fusing sparse touch with pretrained visual representations.
Project page: \href{https://touch-as-saliency.github.io/}{Tac2Pix.github.io}.

\end{abstract}

\section{Introduction}
Dexterous robotic manipulation benefits from complementary sensory modalities. Vision provides dense spatial observations and can leverage powerful pretrained representations for robotic control ~\citep{radosavovic2022realworldrobotlearningmasked,
majumdar2024searchartificialvisualcortex}.
Touch, in contrast, directly measures physical interaction, including contact events and force magnitudes, and remains informative when visual evidence is incomplete or occluded~\citep{Calandra_2018,luo2026touch2robotrobottouchhuman}.
However, tactile sensing is sparse, heterogeneous, and distributed over the robot embodiment, making it difficult to integrate with image-centric visual representations.
Crucially, the spatial structure of robot touch is already known: each tactile measurement is anchored to a specific location on the robot kinematic chain.
Yet most visuo-tactile policies do not explicitly exploit this geometry when relating touch to vision, leaving the correspondence between the two modalities to be learned from limited task demonstrations.

Existing approaches address this mismatch mainly in two ways. Implicit fusion methods encode vision and touch separately and combine their representations through concatenation, modulation, or attention~\citep{feng2024playscorestageguideddynamic,
su2024sim2realmanipulationunknownobjects}.
Although flexible, these methods must infer touch--vision correspondence from task data while operating on sensor-specific tactile representations. A second line of work introduces explicit spatial structure by lifting tactile signals into 3D representations~\citep{wu2025canonicalrepresentationforcebasedpretraining,
huang20253dvitaclearningfinegrainedmanipulation}.
While effective for spatial reasoning, such approaches typically require depth sensing or dedicated 3D processing, making it less direct to reuse the rapidly improving ecosystem of pretrained 2D visual representations.

This motivates a third alternative: grounding touch directly in the native image space of RGB observations.
Rather than requiring the policy to infer cross-modal correspondence from demonstrations through latent-space fusion, we use robot kinematics and camera calibration to establish this correspondence explicitly.
We introduce \textbf{Tac2Pix}, an image-space visuo-tactile fusion framework that converts heterogeneous tactile measurements into a shared, pixel-aligned representation.
Tac2Pix first normalizes tactile sensor responses, then uses forward kinematics and camera calibration to project each measurement onto the RGB image plane, where it is rendered as a force-aware saliency cue.
Together, these steps yield a unified image-space representation of touch that makes its spatial correspondence with vision explicit before feature learning.

By expressing tactile information in the same pixel coordinate system as RGB observations, Tac2Pix can directly leverage the spatial structure and pretrained representations of standard 2D visual backbones.
Inspired by ControlNet~\citep{zhang2023addingconditionalcontroltexttoimage}, we introduce a zero-initialized input adaptation mechanism that adds a tactile saliency channel to the visual encoder while preserving its pretrained RGB pathway at initialization.
Because fusion occurs in image space, the same tactile representation can be used across different visual encoders and downstream policy architectures through lightweight input adaptation.
Tac2Pix thus provides a lightweight, broadly compatible interface between sparse tactile sensing and pretrained visual representations.

We evaluate Tac2Pix on three simulated and three real-world dexterous manipulation tasks, spanning four policy architectures and multiple visual backbones. Tac2Pix achieves strong and consistent performance across these settings, with particularly pronounced gains under degraded visual observations. In real-world experiments, Tac2Pix increases the average success rate under visual occlusion by $26.7$ percentage points over the strongest implicit fusion baseline, with gains persisting under both random and physical occlusions.

Our main contributions are summarized as follows:
\begin{itemize}
    \item We propose \textbf{Tac2Pix}, a tactile-enhanced policy pipeline that converts heterogeneous tactile measurements into force-aware, pixel-aligned saliency maps for policy learning.

    \item We introduce mechanisms for (1) establishing explicit touch--vision correspondence before feature learning, reducing the need to learn this alignment from task demonstrations, and for (2) integrating tactile saliency maps with pretrained 2D visual representations through zero-initialized input adaptation.

    \item We validate Tac2Pix across six dexterous manipulation tasks, four policy architectures, and multiple visual backbones, demonstrating consistent gains in simulation and real-world manipulation, particularly under visual occlusion.
\end{itemize}

\section{related works}
\textbf{Implicit Visuo-Tactile Fusion.}
Tactile sensing provides direct measurements of physical interaction and has been shown to improve robotic manipulation in contact-rich settings, including dexterous manipulation~\citep{guzey2023dexterity,lin2025pp}, regrasping~\citep{Calandra_2018}, slip and contact-state estimation~\citep{dong2018maintaininggraspsslippingbound,she2021cable}, and grasp outcome prediction~\citep{calandra2025feelingsuccessdoestouch}.
As a complementary modality to vision, touch is particularly valuable when visual observations are occluded, degraded, or insufficient for inferring contact states~\citep{10246361,10802719,luo2026blinddexterousgraspingreal2sim2real}.
Most existing visuo-tactile policies combine the two modalities only after separate feature extraction.
For instance, tactile features may be concatenated with visual embeddings~\citep{chen2026dexvitaccollectinghumanvisuotactilekinematic}, used to modulate visual representations through tactile-conditioned parameters~\citep{feng2024playscorestageguideddynamic,Zhu2025TouchIT}, or mixed with visual tokens through attention-based modules~\citep{chen2022visuotactiletransformersmanipulation,pattabiraman2024learningprecisecontactrichmanipulation}.
While flexible, these approaches leave touch--vision spatial correspondence implicit, requiring it to be learned from limited task demonstrations. In contrast, Tac2Pix establishes this correspondence through geometric projection, expressing tactile measurements explicitly in the same pixel coordinate system as RGB and enabling the integration of both modalities within a shared visual encoder.

\textbf{Spatially Grounded Tactile Representations.}
Recent work has increasingly incorporated explicit spatial structure into tactile representations for dexterous manipulation.
Robot Synesthesia~\citep{yuan2024robotsynesthesiainhandmanipulation} represents tactile contacts as point clouds, preserving their 3D spatial organization for visuo-tactile in-hand manipulation. 3DTacDex~\citep{wu2025canonicalrepresentationforcebasedpretraining} introduces a canonical 3D tactile representation that preserves both contact geometry and force information while reducing variation induced by hand configuration.
3D-ViTac~\citep{huang20253dvitaclearningfinegrainedmanipulation} further places visual and dense tactile observations in a unified 3D representation space, explicitly retaining their geometric relationships. Complementarily, SaTA~\citep{huang2026spatiallyanchoredtactileawareness} anchors tactile features to the hand kinematic frame through forward kinematics, providing a stable reference for geometric contact reasoning.
Together, these works demonstrate the importance of preserving spatial structure in tactile perception. However, they represent touch in dedicated 3D, hand-centric, or sensor-layout-aware spaces.
Tac2Pix instead grounds tactile measurements directly in the native coordinate system of RGB observations.
This provides a straightforward interface with existing 2D visual representations, without requiring an additional 3D or hand-centric processing pipeline.

\textbf{Scalable Tactile Representations and Pretrained Visual Backbones.}
Learning general tactile representations remains challenging because tactile sensors vary widely in signal type, spatial layout, sampling characteristics, calibration, and noise profiles~\citep{Yuan2017GelSightHR,bhirangi2024anyskinplugandplayskinsensing,huang2025twintacwiderangehighlysensitive}.
Prior work has learned representations from specific tactile modalities, such as high-resolution vision-based tactile images~\citep{Lambeta_2020,dave2024multimodalvisualtactilerepresentationlearning} and sparse force measurements~\citep{xue2025arraybotreinforcementlearninggeneralizable}, as well as from paired visual-tactile observations for cross-modal alignment~\citep{lee2019makingsensevisiontouch}.
Recent methods further seek reusable tactile representations across sensors and interaction types~\citep{feng2025anytouchlearningunifiedstaticdynamic,rodriguez2024contrastivetouchtotouchpretraining}.
However, unlike RGB images, tactile signals lack a standardized spatial format amenable to large-scale pretraining pipelines, making it difficult to directly reuse the spatial inductive biases and pretrained representations of standard visual backbones~\citep{he2015deepresiduallearningimage,dosovitskiy2021imageworth16x16words}.
Tac2Pix therefore uses image space as a common interface, transforming heterogeneous tactile measurements into pixel-aligned saliency maps that directly leverage the spatial inductive biases and pretrained representations of existing 2D visual backbones.

\section{Methodology}
\begin{figure*}[t]
    \centering
    \includegraphics[width=1\textwidth]{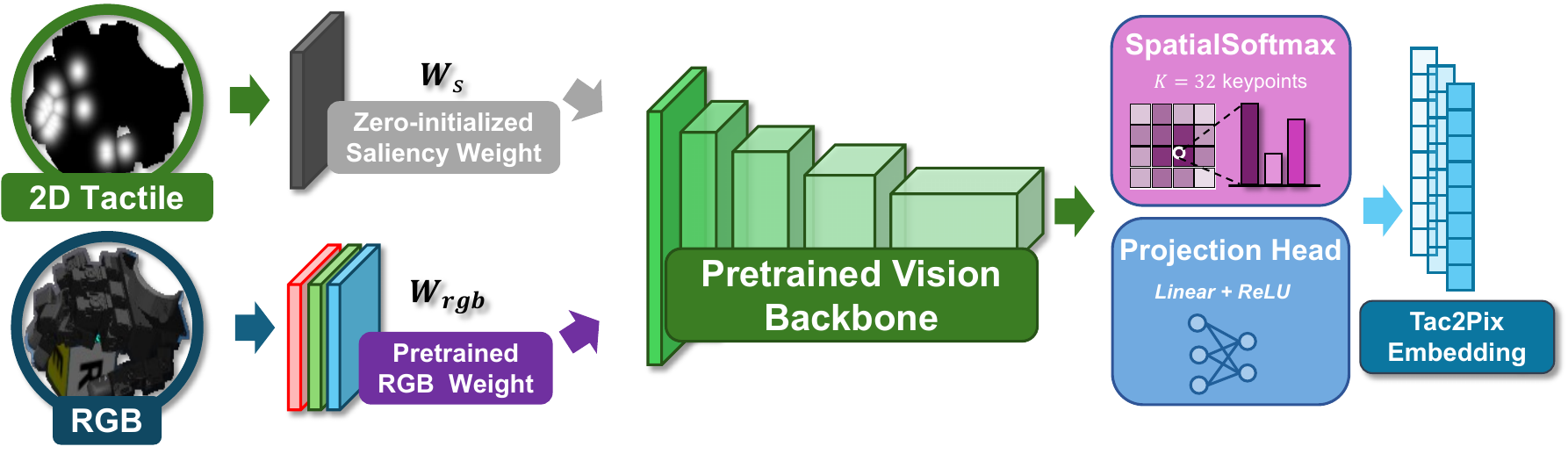}
    \caption{\textbf{The Tac2Pix architecture.} Tac2Pix extends a pretrained RGB visual encoder with a zero-initialized saliency channel, allowing projected tactile cues to be fused in the image domain while preserving the original visual representation at initialization.}
    \label{fig:pipeline}
    \vspace{-10bp}
\end{figure*}

As shown in Figure~\ref{fig:pipeline}, we propose Tac2Pix, a lightweight visuo-tactile policy learning framework that converts sparse, heterogeneous tactile measurements into visually aligned saliency maps.
We first present the problem formulation in Sec.~\ref{subsec:problem_formulation}. We then introduce a visuo-tactile alignment representation in Sec.~\ref{subsec:kinematic_projection}, which maps contact information to image-space saliency cues through the robot kinematic chain and camera projection. Finally, leveraging the proposed representation, we introduce a zero-initialized input adaptation in Sec.~\ref{subsec:rgbs_arch} that integrates tactile saliency into pretrained 2D visual encoders while preserving their original visual pathway at initialization.

\subsection{Problem Formulation}
\label{subsec:problem_formulation}

Our goal is to acquire tactile manipulation skills via imitation learning. We follow the standard visuotactile imitation learning paradigm, in which a policy
$\pi_\theta(\mathbf{a}_t \mid \mathbf{o}_{t-k:t})$
is learned from expert demonstrations
$\mathcal{D} = \{(\mathbf{o}_t, \mathbf{a}_t)\}_{t=1}^{N}$.
Here, $\mathbf{a}_t$ denotes the expert action at time step $t$, and
$\mathbf{o}_t = \{\mathbf{I}_t, \mathbf{s}_t, \mathbf{f}_t\}$
denotes the multimodal robot observation, comprising an RGB image
$\mathbf{I}_t \in \mathbb{R}^{H \times W \times 3}$, robot proprioception
$\mathbf{s}_t$, and low-dimensional tactile measurements $\mathbf{f}_t$.

Unlike approaches that learn cross-modal alignment from latent features, Tac2Pix constructs an aligned visuo-tactile observation explicitly in image space.
Specifically, the raw tactile measurements $\mathbf{f}_t$ are transformed into an image-space saliency map
$\mathbf{S}_t \in \mathbb{R}^{H \times W \times 1}$, which is concatenated with the RGB image to form an augmented RGB-Saliency observation: 
\begin{equation}
    \mathbf{X}_t = \mathrm{Concat}(\mathbf{I}_t, \mathbf{S}_t)
    \in \mathbb{R}^{H \times W \times 4}.
\end{equation}

The learning objective is therefore to train a policy $\pi_\theta(\mathbf{a}_t \mid \mathbf{X}_{t-k:t}, \mathbf{s}_{t-k:t}, \mathbf{f}_{t-k:t}) $
using standard visual imitation learning losses. By embedding tactile information in the same spatial domain as visual observations, this formulation establishes explicit cross-modal spatial correspondences while avoiding ad hoc architectures for heterogeneous tactile modalities. 

\subsection{Force-Aware Tactile-to-Pixel Projection}
\label{subsec:kinematic_projection}

Tac2Pix converts heterogeneous tactile observations into a common image-space representation by addressing two sources of inconsistency: differences in sensor response scale and differences in sensor layout. We first normalize tactile magnitudes across sensor types, and then use robot kinematics and camera geometry to place each tactile response at its corresponding image location.

Specifically, let
\(
\mathbf{f}_t=\{f_{i,t}\}_{i=1}^{M}
\)
denote the tactile readings from $M$ sensing nodes.
Since different tactile sensors may operate at different response scales,
we normalize each reading as
\(
\tilde{f}_{i,t}
=
\tanh\!\left(
\gamma f_{i,t}/F^{i}_{\mathrm{limit}}
\right),
\)
where $F^{i}_{\mathrm{limit}}$ is the sensor-specific saturation limit and
$\gamma$ controls the scaling.
The normalized responses are then placed into a common spatial frame through
kinematic projection.

We next localize each normalized tactile response according to the physical
position of its sensing node.
Each tactile node is associated with a fixed local position $\mathbf{L}_i$ on its corresponding robot link.
Given the robot proprioceptive state $\mathbf{s}_t$, its 3D position in the world frame is obtained through forward kinematics:
\begin{equation}
    \mathbf{P}_{i,t}
    =
    \operatorname{FK}(\mathbf{s}_t,\mathbf{L}_i),
\end{equation}
where $\mathbf{P}_{i,t}\in\mathbb{R}^{3}$.
For camera view $c$, let
$\mathbf{R}^{c}\in\mathbb{R}^{3\times3}$ and
$\mathbf{t}^{c}\in\mathbb{R}^{3}$ denote the camera extrinsics, and
$\mathbf{K}^{c}\in\mathbb{R}^{3\times3}$ the intrinsic matrix.
The corresponding image coordinate
$\mathbf{p}^{c}_{i,t}
=[u^{c}_{i,t},v^{c}_{i,t}]^\top$
is obtained by
\begin{equation}
    [u^{c}_{i,t},v^{c}_{i,t},1]^\top
    \sim
    \mathbf{K}^{c}
    \left(
    \mathbf{R}^{c}\mathbf{P}_{i,t}
    +
    \mathbf{t}^{c}
    \right).
\end{equation}
Projected nodes outside the image bounds are discarded, and the remaining valid nodes form $\mathcal{V}^{c}_{t}$.
Because these projected responses remain spatially sparse, we render them into a dense saliency map $\mathbf{S}^{c}_{t}\in\mathbb{R}^{H\times W\times1}$ using force-aware Gaussian kernels:
\begin{equation}
    \mathbf{S}^{c}_{t}(u,v)
    =
    \max_{i\in\mathcal{V}^{c}_{t}}
    \left[
    \tilde{f}_{i,t}
    \exp\left(
    -\frac{
    (u-u^{c}_{i,t})^{2}
    +
    (v-v^{c}_{i,t})^{2}
    }{
    2\sigma_{i,t}^{2}
    }
    \right)
    \right].
\end{equation}
The spatial support of each tactile response is modulated by its normalized
magnitude:
\begin{equation}
    \sigma_{i,t}
    =
    \sigma_{\min}
    +
    \tilde{f}_{i,t}
    \left(
    \sigma_{\max}-\sigma_{\min}
    \right).
\end{equation}
The tactile magnitude therefore modulates both response strength and spatial support, allowing stronger contacts to exert a broader influence on nearby visual features.
We use max aggregation so that overlapping projected sensors do not artificially amplify the response through summation.

Together, these operations express heterogeneous tactile readings as a common image-space representation for subsequent visuo-tactile learning. The geometric alignment itself is deterministic and contains no learned parameters. The learning process is used purely for developing the
downstream policy that infers action.

\subsection{Zero-Initialized Backbone Adaptation}
\label{subsec:rgbs_arch}

After constructing the observation, we integrate the projected tactile saliency into a pretrained 2D visual encoder.
Rather than introducing a separate tactile encoder, we augment the visual backbone's input with an additional saliency channel. This design preserves the spatial correspondence established in Sec.~\ref{subsec:kinematic_projection} while allowing pretrained RGB visual representations to be reused.

Let $\mathcal{E}_{\mathrm{rgb}}$ denote the pretrained input projection of a visual encoder and $\mathcal{E}_{s}$ the newly introduced projection for the saliency channel. For camera view $c$, the initial visual feature is written as
\begin{equation}
    \mathbf{z}^{c}_{t}
    =
    \mathcal{E}_{\mathrm{rgb}}(\mathbf{I}^{c}_{t})
    +
    \mathcal{E}_{s}(\mathbf{S}^{c}_{t}),
    \qquad
    \mathcal{E}_{s} = \mathbf{0}
    \ \text{at initialization}.
\end{equation}
The pretrained RGB projection is kept unchanged, whereas the parameters associated with the additional saliency input are initialized to zero. Consequently, the Tac2Pix encoder is functionally equivalent to the original pretrained visual encoder at initialization.
During policy learning, the saliency parameters are optimized jointly with the downstream policy, allowing spatially aligned tactile cues to be incorporated without disrupting the pretrained visual pathway at the start of training.

This formulation is independent of a particular visual backbone.
For convolutional encoders, the additional saliency input corresponds to extending the input-channel weights of the first convolutional layer; for patch-based encoders, the same principle is applied to the input patch projection. The resulting per-view visual features are then combined with robot proprioception over the observation horizon to form the conditioning sequence $\mathbf{g}_{t-T_o+1:t}$ used by the downstream policy.

\section{Experiments}
We evaluate the proposed Tac2Pix framework to determine whether image-space tactile grounding improves visuo-tactile imitation learning. We focus on three key research questions:

\hypertarget{rq1}{\textbf{RQ1}}:
Does Tac2Pix consistently improve visuo-tactile policy learning across
tasks, policy architectures, and visual backbones? (Sec.~\ref{subsec:sim_results})

\hypertarget{rq2}{\textbf{RQ2}}:
Is Tac2Pix effective on real-world dexterous manipulation tasks, and how does it compare with alternative spatial tactile representations?
(Sec.~\ref{subsec:real_world_deployment})

\hypertarget{rq3}{\textbf{RQ3}}:
How do saliency rendering, fusion and initialization, and touch--image alignment affect Tac2Pix performance?
(Sec.~\ref{subsec:ablation_alignment})

We conduct experiments in both simulation and real-world environments.
Policies are trained on demonstrations collected under normal, unobstructed visual observations.
They are then evaluated under two conditions: \textit{normal}, with full RGB input, and \textit{occluded}, where a preprogrammed black mask is applied to task-relevant image regions.
The main evaluation considers normal and fixed-occlusion conditions; additional robustness studies further evaluate random and physical occlusions.
The occlusion mask is used only during evaluation, with the same mask size used across tasks.

In simulation, we evaluate three dexterous manipulation tasks: pick-and-place, cube-push, and rotate-cross.
Together, these tasks test object localization, contact-rich interaction, and robustness under partial observability.
For real-world evaluation, we deploy the learned policies on an xArm6 equipped with a LEAP Hand~\citep{shaw2023leap}.
The hand includes 12 joint-mounted FSR sensors and 4 fingertip TwinTac sensors~\citep{huang2025twintacwiderangehighlysensitive}.
Visual observations are captured by two calibrated RealSense D435 cameras.
We evaluate three real-world tasks: pick-and-place, open-drawer, and flip-box, under both normal and occluded observation conditions.
Details on demonstrations, tactile simulation, occlusion masks, and evaluation initialization ranges are provided in Appendix~\ref{app:experimental_details}.

\begin{table*}[t]
\centering
\caption{\textbf{Simulation success rates (\%).} We evaluate multiple visuo-tactile fusion mechanisms across downstream policy architectures and visual conditions. Tac2Pix denotes our image-aligned tactile saliency representation. The best and second-best performances in each column within each policy architecture are highlighted with
\colorbox{purple!18}{lavender}
and
\colorbox{blue!12}{blue}
backgrounds, respectively; ties share the same color.
}
\vspace{-5bp}
\label{tab:sim_success_rate}
\resizebox{\textwidth}{!}{
\begin{tabular}{clccc@{\hspace{6pt}}c
                  ccc@{\hspace{6pt}}c
                  ccc}
\toprule
\multirow{2}{*}[-0.5ex]{\makecell{\textbf{Policy}\\\textbf{Architecture}}} 
& \multirow{2}{*}[-0.5ex]{\makecell{\textbf{Fusion }\\\textbf{Mechanism}}} 
& \multicolumn{3}{c}{Pick-and-Place} & 
& \multicolumn{3}{c}{Cube-Push} & 
& \multicolumn{3}{c}{Rotate-Cross} \\
\cmidrule{3-5} \cmidrule{7-9} \cmidrule{11-13}
& & Normal & Occlud. & Avg. & & Normal & Occlud. & Avg. & & Normal & Occlud. & Avg. \\
\midrule

\multirow{6}{*}{\makecell{Behavior\\Clone MLP\\ \citep{zhang2018deepimitationlearningcomplex}}} 
& Vision-Only & $7.4$ & $0.0$ & $3.7$ & & $ 33.3 $ & $ 0.0 $ & $ 16.7 $ & & $ 10.0 $ & $ 2.0 $ & $ 6.0 $ \\
& Concat & \cellcolor{blue!12}$13.2$ & $0.0$ & \cellcolor{blue!12}$6.6$ & & $ 41.7 $ & $ 0.0 $ & $ 20.8 $ & & \cellcolor{blue!12}$ 28.0 $ & \cellcolor{blue!12}$ 6.0 $ & \cellcolor{blue!12}$ 17.0 $ \\
& FiLM & $6.6$ & $0.0$ & $3.3$ & & \cellcolor{blue!12}$ 50.0 $ & $ 3.3 $ & $ 26.7 $ & & $ 8.0 $ & $ 0.0 $ & $ 4.0 $ \\
& CLiP & $3.3$ & $0.0$ & $1.7$ & & $ 38.3 $ & $ 3.3 $ & $ 20.8 $ & & \cellcolor{blue!12}$ 28.0 $ & $ 2.0 $ & $ 15.0 $ \\
& Cross-Attn& $3.3$ & \cellcolor{purple!18}$8.3$ & $5.8$ & & $ 45.0 $ & \cellcolor{purple!18}$ 21.7 $ & \cellcolor{purple!18}$ 33.3 $ & & $ 18.0 $ & \cellcolor{blue!12}$ 6.0 $ & $ 12.0 $ \\
& \textbf{Ours (Tac2Pix) }
& \cellcolor{purple!18}$15.7$
& \cellcolor{blue!12}$0.8$
& \cellcolor{purple!18}$8.3 $
&  
& \cellcolor{purple!18}$ 53.3 $
& \cellcolor{blue!12}$ 6.7 $
& \cellcolor{blue!12}$ 30.0 $
&  
& \cellcolor{purple!18}$ 38.0 $
& \cellcolor{purple!18}$ 8.0 $
& \cellcolor{purple!18}$ 23.0 $ \\

\midrule
\multirow{6}{*}{\makecell{Action\\Chunk\\Transformer\\ \citep{zhao2023learningfinegrainedbimanualmanipulation}}} 
& Vision-Only & $58.7$ & $0.0$ & $29.3$ & & $ 75.0 $ & $  1.7 $ & $ 38.3 $ & & $ 66.0 $ & $ 16.0 $ & $ 41.0 $ \\
& Concat & \cellcolor{blue!12}$67.8$ & $5.8$ & $36.8$ & & \cellcolor{blue!12}$ 80.0 $ & $ 0.0 $ & $ 40.0 $ & & \cellcolor{purple!18}$ 92.0 $ & $ 20.0 $ & \cellcolor{blue!12}$ 56.0 $ \\
& FiLM & $36.4$ & $0.8$ & $18.6$ & & $ 71.7 $ & $ 13.3 $ & $ 42.5 $ & & $ 60.0 $ & $ 18.0 $ & $ 39.0 $ \\
& CLiP & $38.0$ & \cellcolor{blue!12}$19.0$ & $28.5$ & & $ 60.0 $ & \cellcolor{purple!18}$ 48.3 $ & \cellcolor{blue!12}$ 54.2 $ & & $ 68.0 $ & \cellcolor{blue!12}$ 40.0 $ & $ 54.0 $ \\
& Cross-Attn & \cellcolor{blue!12}$67.8$ & $6.6$ & \cellcolor{blue!12}$37.2$ & & $ 68.3 $ & $ 23.3 $ & $ 45.8 $ & & $ 84.0 $ & $ 26.0 $ & $ 55.0 $ \\
& \textbf{Ours (Tac2Pix)} 
& \cellcolor{purple!18}$71.9$
& \cellcolor{purple!18}$23.1$
& \cellcolor{purple!18}$47.5$
&  
& \cellcolor{purple!18}$ 85.0 $
& \cellcolor{blue!12}$ 46.7 $
& \cellcolor{purple!18}$ 65.8 $
&  
& \cellcolor{blue!12}$ 90.0 $
& \cellcolor{purple!18}$ 44.0 $
& \cellcolor{purple!18}$ 67.0 $ \\

\midrule
\multirow{6}{*}{\makecell{Diffusion\\Policy\\ \citep{chi2024diffusionpolicyvisuomotorpolicy}}} 
& Vision-Only  & $71.9$ & $7.4$ & $39.7$ & & \cellcolor{purple!18}$96.7$ & $ 25.0 $ & $ 60.8 $ & & $ 78.0 $ & $ 26.0 $ & $ 52.0 $ \\
& Concat & \cellcolor{blue!12}$72.7$ & $14.9$ & \cellcolor{blue!12}$43.8$ & & $90.0$ & $ 25.0 $ & $ 57.5 $ & & \cellcolor{blue!12}$ 80.0 $ & $ 38.0 $ & $ 59.0 $ \\
& FiLM & $71.9$ & $13.2$ & $42.6$ & & \cellcolor{blue!12}$95.0$ & $ 38.3 $ & \cellcolor{purple!18}$ 66.7 $ & & $ 64.0 $ & $ 42.0 $ & $ 53.0 $ \\
& CLiP & $62.0$ & $24.8$ & $43.4$ & & $85.0$ & \cellcolor{blue!12}$  43.3$ & \cellcolor{blue!12}$ 64.2 $ & & $ 64.0 $ & \cellcolor{blue!12}$ 48.0 $ & $ 56.0 $ \\
& Cross-Attn & $42.1$ & \cellcolor{blue!12}$34.7$ & $38.4$ & & $71.7$ & \cellcolor{purple!18}$ 46.7 $ & $ 59.2  $ & & $ 70.0 $ & \cellcolor{purple!18}$ 52.0  $ & \cellcolor{blue!12}$ 61.0 $ \\

& \textbf{Ours (Tac2Pix)} 
& \cellcolor{purple!18}$77.7$
& \cellcolor{purple!18}$52.1$
& \cellcolor{purple!18}$64.9$
&  
& $90.0$
& \cellcolor{blue!12}$43.3$ 
& \cellcolor{purple!18}$66.7 $
&  
& \cellcolor{purple!18}$92.0$
& $ 40.0 $
& \cellcolor{purple!18}$ 66.0 $ \\

\midrule
\multirow{6}{*}{\makecell{SmolVLA\\
\citep{shukor2025smolvlavisionlanguageactionmodelaffordable}}}
& Vision-Only
& $62.8$ & $34.7$ & \cellcolor{blue!12}$48.8$ & & \cellcolor{purple!18}$76.7$ & $10.0$ & $43.3$ & & \cellcolor{blue!12}$82.0$ & \cellcolor{blue!12}$64.0$ & $73.0$ \\
& Concat
& \cellcolor{blue!12}$63.6$ & $28.9$ & $46.3$ & & $61.7$ & \cellcolor{blue!12}$30.0$ & $45.8$ & & \cellcolor{purple!18}$84.0$ & \cellcolor{blue!12}$64.0$ & \cellcolor{blue!12}$74.0$ \\
& FiLM
& $57.0$ & $28.1$ & $42.6$ & & $68.3$ & \cellcolor{purple!18}$31.7$ & \cellcolor{blue!12}$50.0$ & & $78.0$ & $60.0$ & $69.0$ \\
& CLiP
& $55.4$ & $32.2$ & $43.8$ & & $68.3$ & $11.7$ & $40.0$ & & \cellcolor{purple!18}$84.0$ & $50.0$ & $67.0$ \\
& Cross-Attn
& $55.4$ & \cellcolor{purple!18}$38.0$ & $46.7$ & & $70.0$ & $26.7$ & $48.3$ & & $72.0$ & \cellcolor{blue!12}$64.0$ & $68.0$ \\
& \textbf{Ours (Tac2Pix)}
& \cellcolor{purple!18}$69.4$ & \cellcolor{blue!12}$36.4$ & \cellcolor{purple!18}$52.9$ & & \cellcolor{blue!12}$71.7$ & \cellcolor{purple!18}$31.7$ & \cellcolor{purple!18}$51.7$ & & \cellcolor{purple!18}$84.0$ & \cellcolor{purple!18}$72.0$ & \cellcolor{purple!18}$78.0$ \\
\bottomrule
\end{tabular}}
\vspace{5px}
\end{table*}

\begin{figure*}[t]
    \centering

    % ================= Left: Learning Curves =================
    \begin{minipage}[t]{0.6\textwidth}
        \vspace{0pt}
        \centering
        \vspace{-10px}
        \includegraphics[width=\linewidth]
        {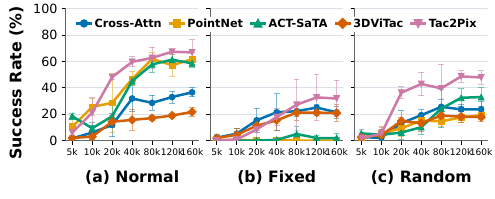}
        \vspace{-25px}
        \caption{\textbf{Learning curves on simulated Pick-and-Place.}
Curves show validation success with standard deviation.}
        \label{fig:learning_curves}
    \end{minipage}
    \hfill
    % ================= Right: Visual Backbones =================
    \begin{minipage}[t]{0.39\textwidth}
        \vspace{0pt}
        \centering
        \small
        \captionof{table}{
        \textbf{Compatibility across visual backbones.}
Using different visual encoders, reported as Normal/Occluded.}
        \label{tab:visual_backbone_compact}

        \setlength{\tabcolsep}{3.5pt}
        \renewcommand{\arraystretch}{1.08}
        \resizebox{\linewidth}{!}{
        \begin{tabular}{lccc}
            \toprule
            \textbf{Method}
            & \textbf{R3M}
            & \textbf{VIP}
            & \textbf{ResNet} \\
            \midrule
            Vision-Only
            & 46.3/9.9
            & 48.8/9.9
            & 58.7/0.0 \\

            Cross-Attn
            & 49.6/4.1
            & 27.3/25.6
            & 67.8/6.6 \\

            \cellcolor{blue!12}\textbf{Tac2Pix}
            & \cellcolor{blue!12}41.3/28.9
            & \cellcolor{blue!12}63.6/27.3
            & \cellcolor{blue!12}71.9/23.1 \\
            \bottomrule
        \end{tabular}}
    \end{minipage}
    \vspace{-10px}
\end{figure*}

\subsection{Simulation Evaluations}
\label{subsec:sim_results}

To answer \hyperlink{rq1}{RQ1}, we evaluate whether Tac2Pix provides consistent benefits across tasks, policy architectures, visual backbones, and visual conditions in simulation.
We consider three manipulation tasks and four policy backbones: Behavior Cloning (BC) MLP~\citep{zhang2018deepimitationlearningcomplex},
Action Chunking Transformer (ACT)~\citep{zhao2023learningfinegrainedbimanualmanipulation},
Diffusion Policy (DP)~\citep{chi2024diffusionpolicyvisuomotorpolicy}, and SmolVLA~\citep{shukor2025smolvlavisionlanguageactionmodelaffordable}.
For each policy class, we compare Tac2Pix with vision-only input,
concatenation, FiLM~\citep{perez2017filmvisualreasoninggeneral},
CLIP-style alignment~\citep{radford2021learningtransferablevisualmodels},
and cross-attention~\citep{jaegle2021perceiver}.

\textbf{Performance across tasks and policies.}
Table~\ref{tab:sim_success_rate} reports success rates for the three simulation tasks under normal and occluded visual conditions. Overall, across tasks and policy architectures, Tac2Pix achieves the best or second-best performance in most settings. We make two main observations. First, simply adding tactile inputs does not consistently improve over vision-only policies.
Concat, FiLM, CLIP-style fusion, and Cross-Attn improve some settings
but degrade others.
This is consistent with recent dexterous-manipulation studies showing
that naive tactile conditioning can introduce irrelevant tactile
interference or degrade existing policy capabilities
~\citep{zhang2026tacvlacontactawaretactilefusion,
niu2026trextactilereactivedexterousmanipulation}.
This suggests that explicitly projecting tactile cues into image space provides a useful geometric prior, avoiding the need to recover touch--vision correspondence implicitly during policy learning.
Second, Tac2Pix shows the strongest overall performance under visual occlusion.
It ranks first or second in most occluded settings and achieves the highest average occluded success across all evaluated fusion mechanisms.
% This suggests that explicitly grounding touch in image space provides a more consistent way to exploit tactile information when visual evidence is incomplete. 

\begin{figure*}[t]
    \centering
    \includegraphics[width=\textwidth]{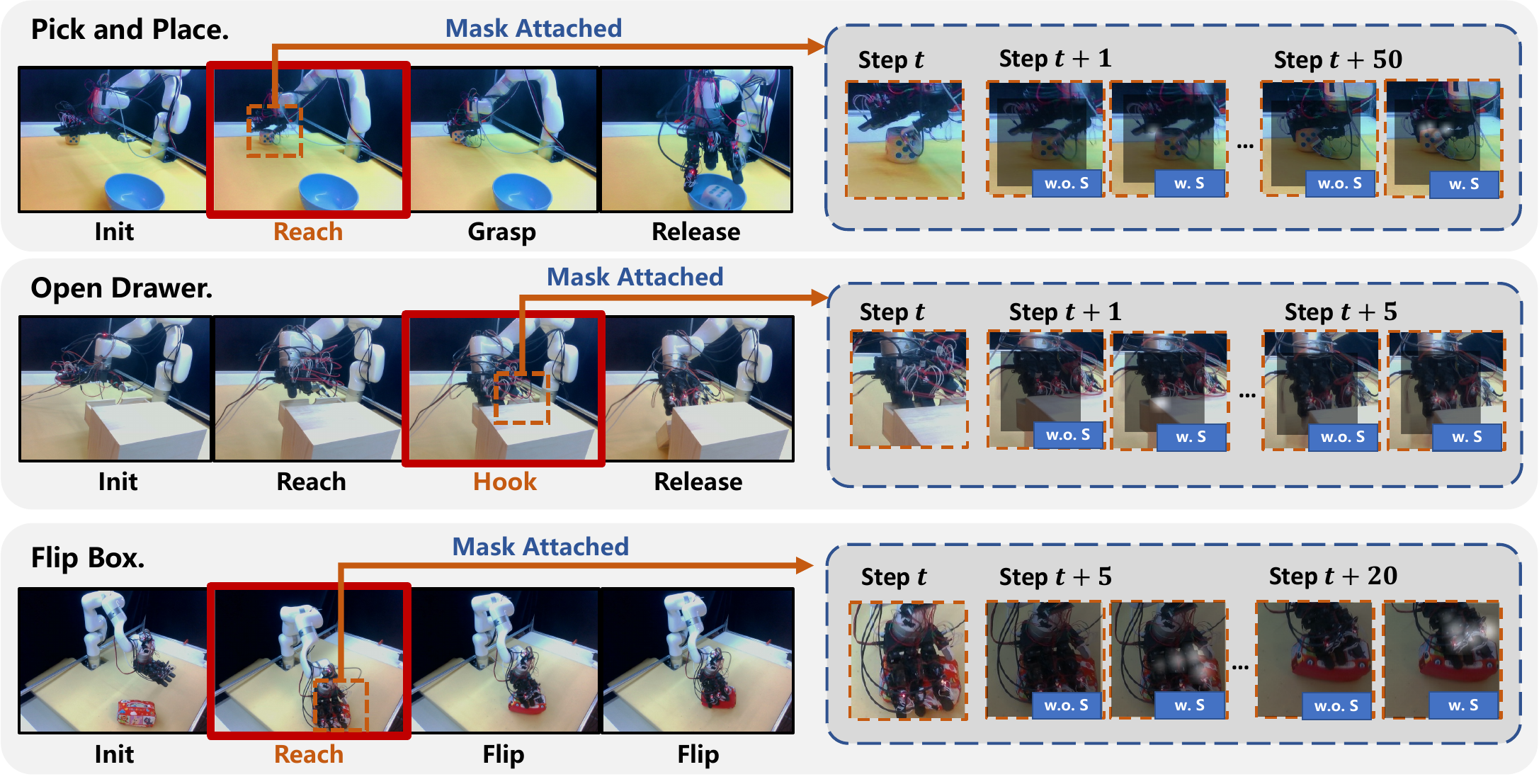}
    \caption{\textbf{Real-world tasks and occlusion protocol.} We evaluate Tac2Pix on three real-world dexterous manipulation tasks by attaching a fixed task-relevant visual mask after the highlighted interaction stage and comparing policy observations with and without tactile saliency. }
    \label{fig:demo}
    \vspace{-15px}
\end{figure*}

Beyond final success rates, Figure~\ref{fig:learning_curves} evaluates simulated Pick-and-Place over three training seeds under normal, fixed, and random occlusion.
Tac2Pix maintains its advantage throughout training and across seeds, while remaining effective when the occlusion pattern changes.
Together, these results indicate that the observed gains are consistent across policy architectures, optimization runs, and visual conditions rather than being tied to a particular initialization or fixed mask.

\textbf{Compatibility across visual backbones.} To isolate the effect of the visual representation, we fix the task to simulated \emph{Pick-and-Place} and use ACT as the downstream policy, while replacing only the visual encoder with R3M~\citep{nair2022r3muniversalvisualrepresentation}, VIP~\citep{ma2023vipuniversalvisualreward}, and ResNet.
As shown in Table~\ref{tab:visual_backbone_compact}, Tac2Pix achieves the highest average success across normal and occluded conditions for all three backbones, as well as the highest success under occlusion in each case.
These results show that the benefit of image-space tactile grounding persists
across the evaluated pretrained visual representations. 
Such compatibility also offers a path for Tac2Pix to benefit from stronger future visual backbones without redesigning the tactile representation.

\subsection{Real-World Deployment}
\label{subsec:real_world_deployment}
To answer \hyperlink{rq2}{RQ2}, we deploy Diffusion Policy on three real-world tasks under normal and visually degraded observations. Figure~\ref{fig:demo} illustrates the task setups, tactile saliency, and the stages at which visual occlusion is introduced.
To assess whether the benefits of image-space tactile grounding transfer across different real-world manipulation behaviors, we compare Tac2Pix with the baselines on all three tasks under both visual conditions.
As shown in Table~\ref{tab:real_world_success}, Tac2Pix achieves the highest average success across \emph{Pick-and-Place}, \emph{Open Drawer}, and \emph{Flip Box}, with substantially larger gains when task-relevant visual
information is occluded.
This suggests that explicit image-space tactile grounding is particularly useful when visual evidence becomes incomplete.

To further test whether the advantage of Tac2Pix extends beyond the fixed-mask setting, we conduct a three-seed real-world evaluation on \emph{Pick-and-Place} and compare against adapted 3D-ViTac and SaTA (Table~\ref{tab:rebuttal}).
We additionally introduce \emph{Random Occ.}, which varies the occluded image region during execution, and \emph{Physical Occ.}, which introduces human-induced occlusion directly in front of the camera.
Tac2Pix achieves the highest success across all four conditions, including normal, fixed, random, and physical occlusion.
Importantly, its advantage persists over both implicit fusion and alternative spatial tactile representations, indicating that directly grounding touch in image space remains effective beyond a particular occlusion pattern or fusion
architecture.

\begin{table*}[t]
    \centering
    \caption{\textbf{Real-world experiments.} We evaluate policies across three tasks under both normal and severely occluded visual conditions. Ours maintains high robustness when vision is compromised.}
    \vspace{-5bp}
    \label{tab:real_world_success}
    \resizebox{\textwidth}{!}{
    \begin{tabular}{lcccccccc}
    \toprule
    \multirow{2}{*}{\textbf{Method}} & \multicolumn{2}{c}{\textbf{Pick \& Place}} & \multicolumn{2}{c}{\textbf{Open Drawer}} & \multicolumn{2}{c}{\textbf{Flip Box}} & \multicolumn{2}{c}{\textbf{Average (\%)}} \\
    \cmidrule(lr){2-3} \cmidrule(lr){4-5} \cmidrule(lr){6-7} \cmidrule(lr){8-9}
    & Normal & Occluded & Normal & Occluded & Normal & Occluded & Normal & Occluded \\
    \midrule
    Vision-Only & $ 9/20 $ & $ 0/20 $ & $ 12/20 $ & $ 4/20 $ & $ 13/20 $ & $ 2/20 $ & $56.7 $ & $ 10.0 $ \\
    Concat & $ 9/20 $ & $ 1/20 $ & $10/20 $ & $ 6/20 $ & $ 14/20 $ & $ 1/20 $ & $55.0 $ & $ 13.3 $ \\
    Cross-Attn & $ 4/20 $ & $ 0/20 $ & $ 6/20 $ & $ 4/20 $ & $ 11/20 $ & $ 11/20 $ & $35.0 $ & $ 25.0 $ \\
    \cellcolor{blue!12}\textbf{Ours (Tac2Pix)} & \cellcolor{blue!12}\textbf{$ 9/20 $} & \cellcolor{blue!12}\textbf{$ 7/20 $} & \cellcolor{blue!12}\textbf{$ 14/20$} & \cellcolor{blue!12}\textbf{$ 10/20 $} & \cellcolor{blue!12}\textbf{$ 17/20 $} & \cellcolor{blue!12}\textbf{$ 14/20  $} & \cellcolor{blue!12}\textbf{$ 66.7 $} & \cellcolor{blue!12}\textbf{$ 51.7 $} \\
    \bottomrule
    \end{tabular}
    }
\end{table*}

\begin{table*}[t]
    \centering
    \caption{\textbf{Real-world robustness beyond fixed visual occlusion.}
Success rates (mean $\pm$ std, \%) over three independently trained seeds
under normal, fixed, random, and physical occlusion.}
    \vspace{-5bp}
    \label{tab:rebuttal}
    \resizebox{\textwidth}{!}{
    \begin{tabular}{lcccc}
        \toprule
        \textbf{Method} & \textbf{Normal} & \textbf{Fixed Occ.}
        & \textbf{Random Occ.} & \textbf{Physical Occ.} \\
        \midrule
        Vision-Only
        & $41.7 \pm 4.7$ & $6.7 \pm 9.4$
        & $23.3 \pm 15.5$ & $25.0 \pm 4.1$ \\
        Cross-Attn
        & $25.0 \pm 4.1$ & $6.7 \pm 4.7$
        & $10.0 \pm 4.1$ & $21.7 \pm 6.2$ \\
        3D-ViTac~\citep{huang20253dvitaclearningfinegrainedmanipulation}
        & $5.0 \pm 4.1$ & $0.0 \pm 0.0$
        & $0.0 \pm 0.0$ & $1.7 \pm 2.4$ \\
        SaTA~\citep{huang2026spatiallyanchoredtactileawareness}
        & $13.3 \pm 2.4$ & $0.0 \pm 0.0$
        & $5.0 \pm 0.0$ & $3.3 \pm 2.4$ \\
        \cellcolor{blue!12}\textbf{Ours (Tac2Pix)}
        & \cellcolor{blue!12}$\mathbf{46.7 \pm 2.4}$
        & \cellcolor{blue!12}$\mathbf{25.0 \pm 7.1}$
        & \cellcolor{blue!12}$\mathbf{28.3 \pm 8.5}$
        & \cellcolor{blue!12}$\mathbf{31.7 \pm 4.7}$ \\
        \bottomrule
    \end{tabular}
    }
    \vspace{-10px}
\end{table*}

% =========================================================
% Saliency Rendering + Initialization/Fusion: side by side
% =========================================================
\begin{table*}[t]
    \centering
    % ---------- Left: Saliency Rendering ----------
    \begin{minipage}[t]{0.49\textwidth}
        \vspace{0pt}
        \centering
        \small
        \captionof{table}{
\textbf{Ablation on tactile saliency rendering.}
We compare RGB overlay, binary contact saliency, and force-aware under both conditions.
}
        \label{tab:ablation_tactile_representation}
        \setlength{\tabcolsep}{4pt}
        \renewcommand{\arraystretch}{1.05}
        \resizebox{\linewidth}{!}{
        \begin{tabular}{lccc}
            \toprule
            \textbf{Variant}
            & \textbf{Normal}
            & \textbf{Occ.}
            & \textbf{Avg.} \\
            \midrule
            Vision-Only
            & 71.9 & 7.4 & 39.7 \\
            RGB Overlay
            & 66.9 & 28.1 & 47.5 \\
            Binary Tac2Pix
            & 70.2 & 44.6 & 57.4 \\
            \cellcolor{blue!12}\textbf{Ours (Force-aware)}
            & \cellcolor{blue!12}\textbf{77.7}
            & \cellcolor{blue!12}\textbf{52.1}
            & \cellcolor{blue!12}\textbf{64.9} \\
            \bottomrule
        \end{tabular}}
    \end{minipage}
    \hfill
    % ---------- Right: Initialization / Fusion ----------
    \begin{minipage}[t]{0.48\textwidth}
        \vspace{0pt}
        \centering
        \small
        \captionof{table}{
\textbf{Ablation on fusion stage and initialization.}
We compare late, intermediate,  and assess the effect of zero initialization.
}
        \label{tab:ablation_architecture}
        \setlength{\tabcolsep}{4pt}
        \renewcommand{\arraystretch}{1.05}
        \resizebox{\linewidth}{!}{
        \begin{tabular}{lccc}
            \toprule
            \textbf{Variant}
            & \textbf{Normal}
            & \textbf{Occ.}
            & \textbf{Avg.} \\
            \midrule
            Late Fusion
            & 64.5 & 14.0 & 39.3 \\
            Intermediate Fusion
            & 69.4 & 26.4 & 47.9 \\
            Early w/o Zero Init.
            & 49.6 & 21.5 & 35.6 \\
            \cellcolor{blue!12}\textbf{Ours (Early)}
            & \cellcolor{blue!12}\textbf{77.7}
            & \cellcolor{blue!12}\textbf{52.1}
            & \cellcolor{blue!12}\textbf{64.9} \\
            \bottomrule
        \end{tabular}}
    \end{minipage}
    \vspace{-0.05cm}
\end{table*}

\subsection{Ablation on Tac2Pix Design Choices}
\label{subsec:ablation_alignment}

To answer \hyperlink{rq3}{RQ3}, we ablate three key design choices in Tac2Pix: 
(1) the tactile saliency rendering strategy, to examine whether alternative rendering variants yield better performance; 
(2) the visuo-tactile fusion architecture, to assess the effectiveness of the proposed fusion mechanism; and 
(3)  spatial alignment between visual and tactile inputs, to evaluate robustness to projection errors. 
We conduct all ablations using Diffusion Policy on the pick-and-place task and compare success rates across design variants.

\textbf{Ablation on Tac2Pix rendering variants.}
Our proposed force-aware Tac2Pix represents projected contacts as force-modulated Gaussian kernels. To validate this rendering design, we compare it with two alternatives in simulation: 
(a) \textit{RGB Overlay}, which renders tactile cues directly on the RGB image while preserving a standard three-channel input; and 
(b) \textit{Binary Tac2Pix}, which appends a fourth saliency channel but assigns a constant intensity to all active contacts, thereby encoding contact location without force magnitude.

Table~\ref{tab:ablation_tactile_representation} summarizes the results. Under normal, unobstructed vision, the performance differences among the representations are relatively minor. Under occlusion, however, these differences become more pronounced: both tactile-augmented variants substantially outperform the vision-only baseline. In particular, binary Tac2Pix demonstrates that encoding contact location alone already provides a strong geometric prior for manipulation. Nevertheless, force-aware Tac2Pix achieves the highest success rates in both visual settings, suggesting that continuous force magnitude conveys additional interaction information beyond binary contact.

\textbf{Ablation on fusion architecture.}
We examine the effects of fusion stage and initialization through three variants:
(1) \textit{Late Fusion}, which encodes RGB and saliency separately before pooling and concatenating their features;
(2) \textit{Intermediate Fusion}, which encodes saliency with a lightweight convolutional network and adds its features to an intermediate ResNet feature map; and
(3) \textit{Early Fusion w/o Zero Initialization}, which retains our architecture but uses standard initialization for the added saliency-channel weights.

As shown in Table~\ref{tab:ablation_architecture}, intermediate fusion outperforms late fusion under both visual conditions, with a larger advantage under occlusion. Removing zero initialization reduces early-fusion success under normal vision and  occlusion. The full model achieves the highest success in both settings, supporting the combination of early tactile injection and zero-initialized input adaptation.

\textbf{Ablation on spatial alignment.} 
Finally, to assess the importance of touch--image alignment, Table~\ref{tab:projection_noise} quantifies Tac2Pix's sensitivity to spatial misalignment. To this end, we introduce controlled pixel shifts $(\Delta_x, \Delta_y)$ to tactile saliency map. The offset is fixed within each trial and randomly sampled across trials: for saliency map $S\in\mathbb{R}^{H\times W}$, we construct
\begin{equation}
S_{\Delta}(x,y)=S\big((x-\Delta_x)\bmod W,\,(y-\Delta_y)\bmod H\big),
\end{equation}
while leaving the RGB image unchanged. The results show that performance generally decreases as spatial misalignment increases, with a larger drop under visual occlusion than under normal vision. The same qualitative trend appears in simulation and the real world, confirming that correct touch--image correspondence becomes particularly important when visual evidence is degraded.

\begin{table*}[t]
    \centering
    \caption{\textbf{Sensitivity to spatial misalignment.}
Success under controlled translational offsets of the tactile saliency map while keeping RGB unchanged, evaluated in simulation and the real world.}
    \label{tab:projection_noise}
    \renewcommand{\arraystretch}{1.08}
    \setlength{\tabcolsep}{14pt}

    \begin{tabular}{llcccc}
        \toprule
        \textbf{Setting}
        & \textbf{Condition}
        & \textbf{0 px}
        & \textbf{25 px}
        & \textbf{50 px}
        & \textbf{100 px} \\
        \midrule
        \multirow{2}{*}{Sim}
        & Normal & 77.7\% & 71.9\% & 71.1\% & 67.8\% \\
        & Occluded   & 52.1\% & 27.3\% & 31.4\% & 18.2\% \\
        \midrule
        \multirow{2}{*}{Real}
        & Normal & 9/20 & 8/20 & 5/20 & 5/20 \\
        & Occluded   & 7/20 & 4/20 & 3/20 & 3/20 \\
        \bottomrule
    \end{tabular}
    \vspace{-10px}
\end{table*}

Overall, the experiments support a reusable design principle: known contact geometry can be expressed in a visual form that standard policy models can consume directly.
Tac2Pix shows that this representation yields practical benefits in both
simulation and real-world manipulation, particularly when vision is unreliable.

\section{Conclusion}
In this paper, we introduced Tac2Pix, a visuo-tactile framework that projects tactile contacts onto the 2D image plane as force-aware saliency maps.
Through zero-initialized input adaptation, Tac2Pix injects explicit spatial priors into standard visual encoders while preserving pretrained visual representations.
Across simulated and real-world experiments, Tac2Pix consistently improves visuo-tactile imitation learning, particularly when vision is degraded.
These results support image space as a simple, broadly compatible interface for integrating sparse touch with visual representations.
Tac2Pix currently relies on known sensor geometry and calibrated camera parameters, and its 2D projection does not preserve full depth information.
Future work will explore calibration-robust projection.
More broadly, Tac2Pix offers a path to connect touch with the rapidly evolving visual learning community, enabling robot policies to leverage advances in visual backbones without large-scale tactile pretraining.

% In this paper, we introduced Tac2Pix, a visuo-tactile framework that bridges cross-modal asymmetry by projecting tactile contacts onto the 2D image plane as force-modulated saliency maps. Coupled with a zero-initialized conditioning mechanism, Tac2Pix embeds explicit spatial priors directly into standard visual encoders while preserving the integrity of pretrained representations. Across simulation and real-world experiments, Tac2Pix improves vision-based imitation learning, especially under visual occlusion. These results show that representing touch as image-aligned saliency is a simple, policy-agnostic, and effective approach for improving robustness in contact-rich manipulation.

\newpage

\subsection*{acknowledgments}
{This work was supported by the National Natural Science Foundation of China (Grant No. 52305007), the Natural Science Foundation of Shanghai (Grant No. 25ZR1402370), the Artificial Intelligence Project of the State Key Laboratory of General Artificial Intelligence, BIGAI, Peking University, Beijing, China (Project No. SKLAGI2025OP19), the State Key Laboratory of Mechanical System and Vibration (Grant No. MSV202519) and the MoE Key Laboratory of Intelligent Perception and Human-Machine Collaboration (KLIP-HuMaCo).}

\subsection*{AI use statement}

% (This section is \textbf{required} and does not count toward the page limit.)

In this work, we used generative AI tools to search for and summarize relevant literature, provide feedback on research methods and experimental design, suggest experimental parameters, write and edit software code, clean and reformat data, support data analysis, and improve the readability of the manuscript. We did not use generative AI tools to develop theoretical models or conceptual frameworks, propose or refine hypotheses, generate synthetic datasets, formulate mathematical claims, contribute essential steps to mathematical proofs, write proofs, or interpret the results. Translation was not applicable to this work. The authors reviewed all AI-assisted work and verified and tested AI-generated code. We take responsibility for the final content of this work, including any text, claims, or artifacts produced with the assistance of generative AI.

% See the ICLR 2027 AI Policy for Authors for more details. This statement should
% not be more than 1 page.

% \subsection*{Reproducibility statement}

% (This section is \textbf{recommended} and does not count toward the page limit.)

% It is important that the work published in ICLR is reproducible. Authors are
% strongly encouraged to include a paragraph-long Reproducibility Statement at the
% end of the main text (before references) to discuss the efforts that have been
% made to ensure reproducibility. This paragraph should not itself describe
% details needed for reproducing the results, but rather reference the parts of
% the main paper, appendix, and supplemental materials that will help with
% reproducibility. For example, for novel models or algorithms, a link to an
% anonymous downloadable source code can be submitted as supplementary materials;
% for theoretical results, clear explanations of any assumptions and a complete
% proof of the claims can be included in the appendix; for any datasets used in
% the experiments, a complete description of the data processing steps can be
% provided in the supplementary materials. Each of the above are examples of
% things that can be referenced in the reproducibility statement.

\subsection*{Reproducibility statement}

We are committed to ensuring the reproducibility of our work. We provide detailed experimental protocols, baseline adaptations, ablation settings, and policy/training configurations in the appendix
(Secs.~\ref{app:experimental_details}--\ref{app:downstream_policies}).
Our data generation, training, and inference pipelines are built on the open-source LeRobot framework. We plan to release the modular Tac2Pix implementation together with the configurations to reproduce our experiments upon acceptance.

\bibliography{iclr2027_conference}
\bibliographystyle{iclr2027_conference}

\newpage
\appendix
\section*{Appendix Overview}
\label{app:overview}

This appendix provides additional details on the experimental setup, simulation and real-world task configurations, qualitative analyses, fusion mechanisms, and downstream policy training.

\vspace{0.5em}
\noindent
\textbf{\hyperref[app:experimental_details]{A. Experimental Setup and Task Details}}  
describes the real-world hardware platform, teleoperation interface, observation and action spaces, tactile sensing layout, and demonstration collection protocol. 
It also details the simulation and real-world task configurations, including saliency generation, task initialization ranges, and demonstration statistics.

\vspace{0.35em}
\noindent
\textbf{\hyperref[app:extended_discussions]{B. Extended Discussions and Visualizations}}  
provides additional analyses of the proposed tactile saliency representation, including depth ambiguity in 2D projection, inference efficiency compared with 3D baselines, and Grad-CAM visualizations under visual occlusion.

\vspace{0.35em}
\noindent
\textbf{\hyperref[app:saliency_ablations]{C. Saliency Fusion Ablation Details}}  
describes details of different saliency fusion architectures mentioned in Sec~\ref{subsec:ablation_alignment}.

\vspace{0.35em}
\noindent
\textbf{\hyperref[app:fuse_mechms]{D. Training Details of Fusion Mechanisms}}  
describes the implementation details of the evaluated visuo--tactile fusion baselines, including cross-attention fusion, FiLM modulation, and CLIP-based vision--tactile contrastive pretraining.

\vspace{0.35em}
\noindent
\textbf{\hyperref[app:downstream_policies]{E. Training Details of Downstream Policies}}  
summarizes the downstream policy backbones used in our experiments, including BC-MLP, ACT, and Diffusion Policy, together with their architectures, training objectives, and hyperparameters.

\vspace{1em}

\section{Experimental Setup and Task Details}
\label{app:experimental_details}

\begin{figure*}[h]
    \centering
    \includegraphics[width=\textwidth]{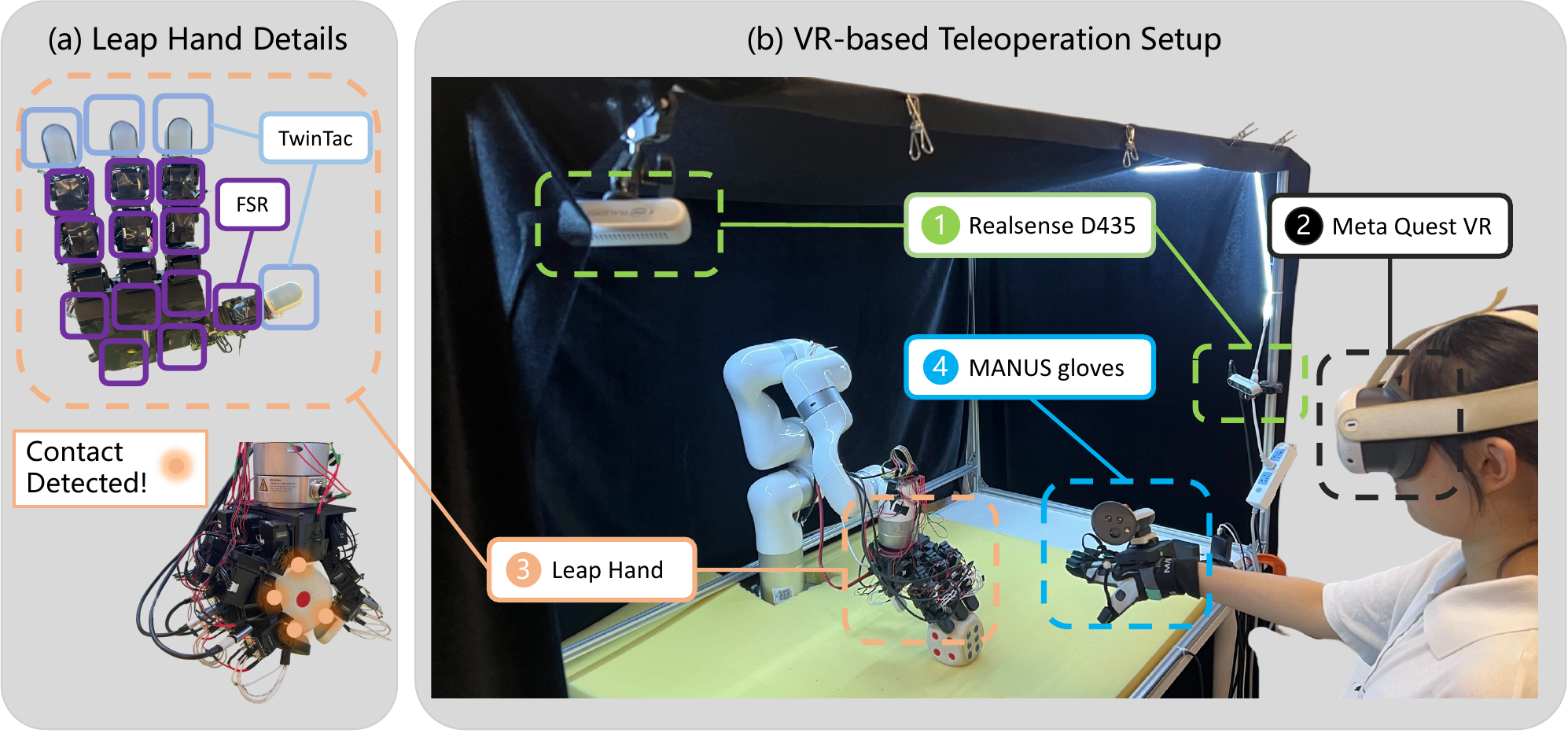}
    \caption{
    \textbf{Real-world teleoperation and deployment platform.}
    }
    \label{fig:hardware_setup}
    \vspace{-15bp}
\end{figure*}

The complete physical setup, including the tactile sensor layout and camera viewpoints, is shown in Fig.~\ref{fig:hardware_setup}. 
We use a unified teleoperation interface for both simulated and real-world data collection. 
The operator controls the xArm6 and LEAP Hand using a Meta Quest 3 headset for wrist tracking and visual feedback, together with a Manus Quantum Metaglove for finger-motion capture. 
Motion retargeting and data recording are built on the open-source ARCap framework~\citep{chen2024arcapcollectinghighqualityhuman}. 
For each task, we collect task-successful expert demonstrations using visual teleoperation, while robot tactile signals are synchronously recorded during execution. 
Although the operator does not receive tactile or haptic feedback, the recorded tactile observations capture the robot's contact state along the expert trajectory and are used as additional policy inputs during learning.
Unless otherwise stated, evaluation uses a separate set of initial states that expands the demonstration initialization range to test robustness beyond the training distribution.

All policies use the same observation and action interface in simulation and the real world. 
At each timestep, the observation contains proprioception, two RGB camera views, and tactile readings. 
The proprioceptive state is
$
    s_t = [q^{\mathrm{arm}}_t, q^{\mathrm{hand}}_t] \in \mathbb{R}^{22},
$
including 6 arm joints and 16 hand joints. 
Each RGB image is center-cropped and resized to $224 \times 224$. 
For Tac2Pix policies, we additionally provide one saliency map per camera view at the same resolution. 
Unless otherwise stated, all models are trained on NVIDIA A40 GPUs and deployed on an NVIDIA RTX 4090 GPU.

The tactile observation consists of $32$ taxel readings from four TwinTac fingertip sensors~\citep{huang2025twintacwiderangehighlysensitive} and $12$ FSR readings distributed across the hand:
$
    \tau_t = [\tau^{\mathrm{taxel}}_t, \tau^{\mathrm{fsr}}_t] \in \mathbb{R}^{44},
$
where $\tau^{\mathrm{taxel}}_t \in \mathbb{R}^{32}$ and $\tau^{\mathrm{fsr}}_t \in \mathbb{R}^{12}$. 
The policy outputs a 22-dimensional target command,
$
    a_t = [a^{\mathrm{arm}}_t, a^{\mathrm{hand}}_t] \in \mathbb{R}^{22}.
$
All demonstrations and policy rollouts are represented at 20 Hz.

% [Please supplement further data collection details here...]}

\subsection{Simulation Task Details}
\label{app:task_simulation}
\paragraph{Saliency generation in simulation.}
In simulation, tactile signals are generated using a tactile simulator built on ETac~\citep{xu2026etac}. 
Simulation uses the same 44-node tactile layout as the real robot, consisting of 32 TwinTac taxels and 12 FSR nodes.
Each node is attached to its corresponding hand link with a fixed local offset, and its 3D position is obtained through forward kinematics from the simulated joint state.
The resulting 44 scalar tactile readings are provided identically to all
methods. For Tac2Pix, these same nodes are projected into each camera view and rendered using the force-dependent Gaussian formulation in Sec.~\ref{subsec:kinematic_projection}. We use $\sigma_{\min}=4$ and $\sigma_{\max}=12$ in all experiments.

\paragraph{Pick and Place.}
The Pick-and-Place task evaluates whether a policy can grasp a cube-like object, lift it from the tabletop, and drop it into a target region.
We collect 53 expert demonstrations, totaling 20,212 frames.
During demonstration collection, the cube's initial position is uniformly sampled from a $12\times 12~\mathrm{cm}$ region, as shown in Fig.~\ref{fig:init_workspace}d.
For evaluation, we test policies on 121 initial states, including both in-distribution and out-of-distribution (OOD) cases; the OOD split expands the sampling region to $16\times 16~\mathrm{cm}$.

\paragraph{Cube Push.}
The Cube-Push task evaluates whether a policy can perform planar, contact-rich manipulation by pushing a cube into a target slot.
We collect 32 expert demonstrations, totaling 9,500 frames.
During demonstration collection, the cube is initialized uniformly within a $2\times 8~\mathrm{cm}$ area on the tabletop, as shown in Fig.~\ref{fig:init_workspace}e.
For evaluation, we test policies on 60 initial configurations, with cube positions uniformly sampled from a $2\times 16~\mathrm{cm}$ area.

\paragraph{Rotate Cross.}
The Rotate-Cross task evaluates whether a policy can rotate a valve-like object through sustained contact.
We collect 25 expert demonstrations, totaling 11,084 frames.
During demonstration collection, the cross is initialized uniformly within a $10\times 10~\mathrm{cm}$ area on the tabletop, as shown in Fig.~\ref{fig:init_workspace}f.
For evaluation, we test policies on 50 initial configurations, with cross positions uniformly sampled from a $12\times 12~\mathrm{cm}$ area.

\subsection{Real Tasks Details}
\label{app:task_real}

\paragraph{Real-world saliency generation.}
Real-world policies use the observation and action definition described in Sec.~\ref{app:experimental_details}. 
For Tac2Pix policies, real-world saliency is computed from the physical tactile sensing nodes on the hand: $4 \times 8$ TwinTac fingertip taxels and $12$ FSR sensors, yielding $44$ projected tactile nodes in total.
At each timestep, the 3D position of each tactile node is computed from the robot forward kinematics and its fixed local offset on the corresponding hand link. 
We calibrate the camera extrinsics using EasyHEC~\citep{Chen_2023} and use the calibrated RealSense intrinsics for image projection. 
The projected tactile nodes are then rendered into image-space saliency maps using the force normalization, Gaussian rendering, and max-aggregation procedure described in Sec.~\ref{subsec:kinematic_projection}. 
This produces one saliency map for each camera view.

\begin{figure*}[t]
    \centering
    \includegraphics[width=\textwidth]{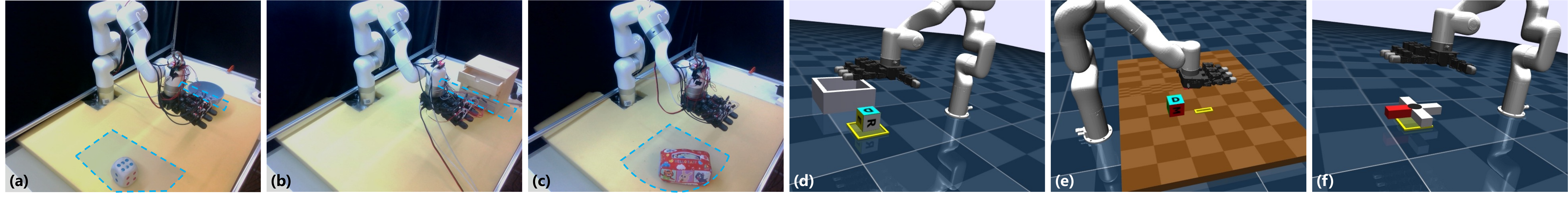}
    \caption{
    \textbf{Initialization workspace of operational objects.} The blue areas in real tasks and yellow areas in simulation tasks are initial workspace of operation objects. (a) Pick and place (Real). It is noteworthy that the bowl  where the cube should be lifted in, is also randomly placed initially; (b) Open drawer. (Real); (c) Flip box(Real); (d) Pick and place (Sim); (e) Cube Push (Sim); (f) Rotate cross (Sim).
}
    \label{fig:init_workspace}
    \vspace{-10bp}
\end{figure*}

\paragraph{Pick and Place.}
The task configuration is similar to that in simulation. 
The policy must grasp a randomly initialized cube on one side of the workspace and drop it into a specified bowl on the other side.
We collect 48 expert demonstrations for this task, corresponding to 23,820 frames, with initializations shown in Fig.~\ref{fig:init_workspace}a.
During demonstration collection, the cube is initialized randomly within a $20\times 30~\mathrm{cm}$ area on the tabletop.

\paragraph{Open Drawer.}
The Open-Drawer task evaluates whether the dexterous hand can hook its fingers onto the edge of a partially opened drawer and pull it outward to complete the drawer-opening motion.
We collect 48 expert demonstrations for this task, corresponding to 11,848 frames, with initializations shown in Fig.~\ref{fig:init_workspace}b.
During demonstration collection, the drawer pose is initialized randomly within a $5\times 30~\mathrm{cm}$ area on the tabletop.

\paragraph{Flip Box.}
The Flip-Box task evaluates whether the dexterous hand can use its fingers to manipulate a tissue box twice along its long edge, rotating it by a total of 180 degrees until its bottom face points upward.
We collect 26 expert demonstrations for this task, corresponding to 12,125 frames, with initializations shown in Fig.~\ref{fig:init_workspace}c.
During demonstration collection, the tissue box is initialized within a sector-shaped region with a radius of $35~\mathrm{cm}$ on one side of the tabletop.

\paragraph{Relation between the two real-world evaluations.}
Table~\ref{tab:real_world_success} reports the original single-seed
evaluation across all three real-world tasks, with 20 trials per task
and condition. Table~\ref{tab:rebuttal} expands the Pick-and-Place
evaluation to three independently trained seeds and additionally includes
random and physical occlusion settings. The original Pick-and-Place run
is included as one of the three seeds in the expanded evaluation.

\paragraph{Occlusion protocols.}
We evaluate three visual-degradation settings.
\textit{Fixed Occlusion} uses the task-specific software mask from the
main evaluation, with the mask activated after the predefined interaction
stage and kept fixed for the remainder of the rollout.
\textit{Random Occlusion} uses the same mask size and activation stage,
but relocates the mask every 20 frames among predefined task-relevant
image regions.
The complete mask schedule is shared across methods for matched evaluation.
\textit{Physical Occlusion} introduces dynamic human-induced occlusion
directly in front of the camera, without digitally modifying the RGB stream.

\paragraph{Adaptation of spatial tactile baselines.}
We adapt 3D-ViTac
by replacing its dense tactile arrays with our 44 sparse sensing nodes.
Each node is represented by its tactile reading and its 3D position
computed through forward kinematics.
We combine these tactile points with 512 points from camera depth images,
giving priority to points near the tactile sensors and within the
interaction workspace.
Following the original framework, we use an adapted PointNet++ encoder
to process the combined points and a diffusion policy to predict actions.

We replace SaTA's tactile-image encoder with a shared MLP over the eight taxel readings from each TwinTac fingertip.
We retain spatial anchoring through Fourier encoding and FiLM, using fingertip poses computed by forward kinematics in the palm frame.
The 12 FSR readings form an additional global token, which joins the four fingertip tokens, RGB features, and proprioception in the ACT policy.
The lower performance of the adapted 3D-ViTac and SaTA baselines likely reflects a sensing–representation mismatch: their original formulations exploit richer spatial tactile observations, whereas our setting provides sparse scalar readings.

\section{Extended Discussions and Visualizations}
\label{app:extended_discussions}

In this section, we provide additional analyses and qualitative visualizations to discuss depth ambiguity in our projection model (Sec.~\ref{app:depth_ambiguity}), the efficiency analysis (Sec.~\ref{app:efficiency}), and the policy's attention mechanism under visual occlusion  (Sec.~\ref{app:attention_maps}).

\subsection{Addressing Depth Ambiguity in 2D Projection}
\label{app:depth_ambiguity}

A limitation of our kinematic projection model is its limited expressivity in representing the full 3D contact location. In particular, contacts occurring on the side of the object facing away from the camera may still project onto the 2D foreground. This creates depth ambiguity: from a single 2D saliency map, the model cannot determine whether a contact lies on the front or rear surface of the object, nor can it precisely distinguish which front or rear tactile sensor is activated.

However, this ambiguity has little effect on overall performance across the tasks we evaluate, as reflected in Tables~\ref{tab:sim_success_rate} and~\ref{tab:real_world_success}. We attribute this robustness to three factors. First, the policy has access to the 3D kinematic configuration of the robot fingers through proprioception, which provides information beyond the projected saliency map. Second, visual observations are aggregated from multiple camera viewpoints, reducing reliance on any single ambiguous projection. Third, our network is robust to contact offsets, as analyzed in Table~\ref{tab:projection_noise}. As a result, the network is not required to infer the complete 3D contact state from a single 2D view alone, and the limited expressivity of the projection model does not significantly degrade performance.

{
\subsection{Computational Efficiency}
\label{app:efficiency}

\begin{table}[t]
    \centering
    \footnotesize
    \setlength{\tabcolsep}{3.0pt}
    \renewcommand{\arraystretch}{1.15}
    \caption{\textbf{Efficiency comparison per step.} Denoising, pre-denoising, and overall time costs are reported in milliseconds.}
    \label{tab:efficiency_latency}
    \begin{tabular}{@{}>{\raggedright\arraybackslash}m{0.20\linewidth}ccc@{}}
        \toprule
        \textbf{Model} &
        \makecell[c]{\textbf{Denoising}\\\textbf{Time (ms)}} &
        \makecell[c]{\textbf{Pre-denoising}\\\textbf{Latency (ms)}} &
        \makecell[c]{\textbf{Overall}\\\textbf{Time (ms)}} \\
        \midrule
        Vision-Only  & $64.26 \pm 0.01$ & $10.10 \pm 5.46$ & $74.36$ \\
        Concat       & $64.23 \pm 0.00$ & $10.23 \pm 4.24$ & $74.46$ \\
        FiLM         & $64.24 \pm 0.00$ & $11.60 \pm 3.31$ & $75.84$ \\
        CLiP         & $66.39 \pm 0.00$ & $7.81 \pm 2.01$  & $74.20$ \\
        Point Cloud  & $76.72 \pm 0.00$ & $95.12 \pm 7.83$ & $171.84$ \\
        Cross-Attn   & $64.56 \pm 0.00$ & $15.13 \pm 3.96$ & $79.69$ \\
        Ours (Tac2Pix) & $64.24 \pm 0.00$ & $21.06 \pm 4.54$ & $85.30$ \\
        \bottomrule
    \end{tabular}
\vspace{-10bp}
\end{table}

For real-time deployment, low inference latency is essential for high-frequency closed-loop control. We compare Tac2Pix with the baselines across different inference stages. Beyond comparing visuo-tactile fusion mechanisms, this analysis contextualizes the efficiency of image-space tactile grounding against an alternative explicit 3D grounding strategy.

Tac2Pix projects tactile contacts onto the 2D image plane and reuses a standard visual encoder. In contrast, a natural alternative is to introduce an explicit 3D geometric branch, where depth-derived point-cloud features are encoded separately and fused with RGB and tactile representations at the latent level. While this provides additional 3D spatial context, it typically requires depth input, point-cloud preprocessing, and a separate 3D feature extractor, which can introduce non-negligible latency in closed-loop deployment. As a reference, in addition to the baselines used in the main experiments, we include a 3D visuo-tactile policy following~\citep{wu2023learning}. This policy uses RGB, depth, and tactile inputs by concatenating embeddings from pretrained modality-specific encoders. Specifically, it uses ResNet-18 for image features and PointNet~\citep{qi2017pointnetdeeplearningpoint} for point-cloud features.

The efficiency comparison results are shown in Tab.~\ref{tab:efficiency_latency}, we first measure the denoising-stage inference cost after the global condition has been constructed. In this setting, the main difference is the conditional vector passed to the diffusion model. Most policies require about $64$--$66$\,ms per control step. However, the 3D policy has a substantially longer conditioning vector due to the concatenated PointNet features, increasing its inference time to $76.72$\,ms. In contrast, Tac2Pix and the vision-only baseline require only $64.24$\,ms and $64.26$\,ms, respectively.

We also report pre-processing latency in Tab.~\ref{tab:efficiency_latency}, including all preprocessing and feature extraction before diffusion denoising. The 3D policy incurs the largest preprocessing cost, $95.12 \pm 7.83$\,ms, dominated by point-cloud processing: farthest point sampling and PointNet feature extraction together take $68.23$\,ms. Tac2Pix is much faster, with a preprocessing latency of $21.06 \pm 4.54$\,ms; saliency generation itself takes only $6.14 \pm 1.89$\,ms. These results show that forward-kinematics-based saliency computation is lightweight enough for real-time deployment and does not introduce a prohibitive latency burden.

Overall, Tac2Pix offers a favorable efficiency--robustness trade-off. It preserves the denoising-stage speed of standard 2D visual diffusion policies while adding only modest preprocessing overhead for tactile saliency construction.

\subsection{Feature Attention Visualizations Under Occlusion}
\label{app:attention_maps}

To illustrate how the policy uses tactile cues under visual occlusion, we visualize feature attention with Grad-CAM~\citep{selvaraju2017grad} in Fig.~\ref{fig:sim_gradcam}. 
Red indicates stronger model attention, while blue indicates weaker attention. 
For each task, we compare Grad-CAM maps across models under normal and occluded settings.

Under normal visibility, highlighted regions typically appear around the object and the robot hand. 
Under occlusion, baseline models tend to shift attention away from task-relevant regions. 
In contrast, Tac2Pix continues to focus near the hand, suggesting that it learns to rely on projected tactile cues when visual cues are unreliable.

\begin{figure}[t]
    \centering
    \includegraphics[
        width=0.95\linewidth,
        height=0.9\textheight,
        keepaspectratio,
    ]{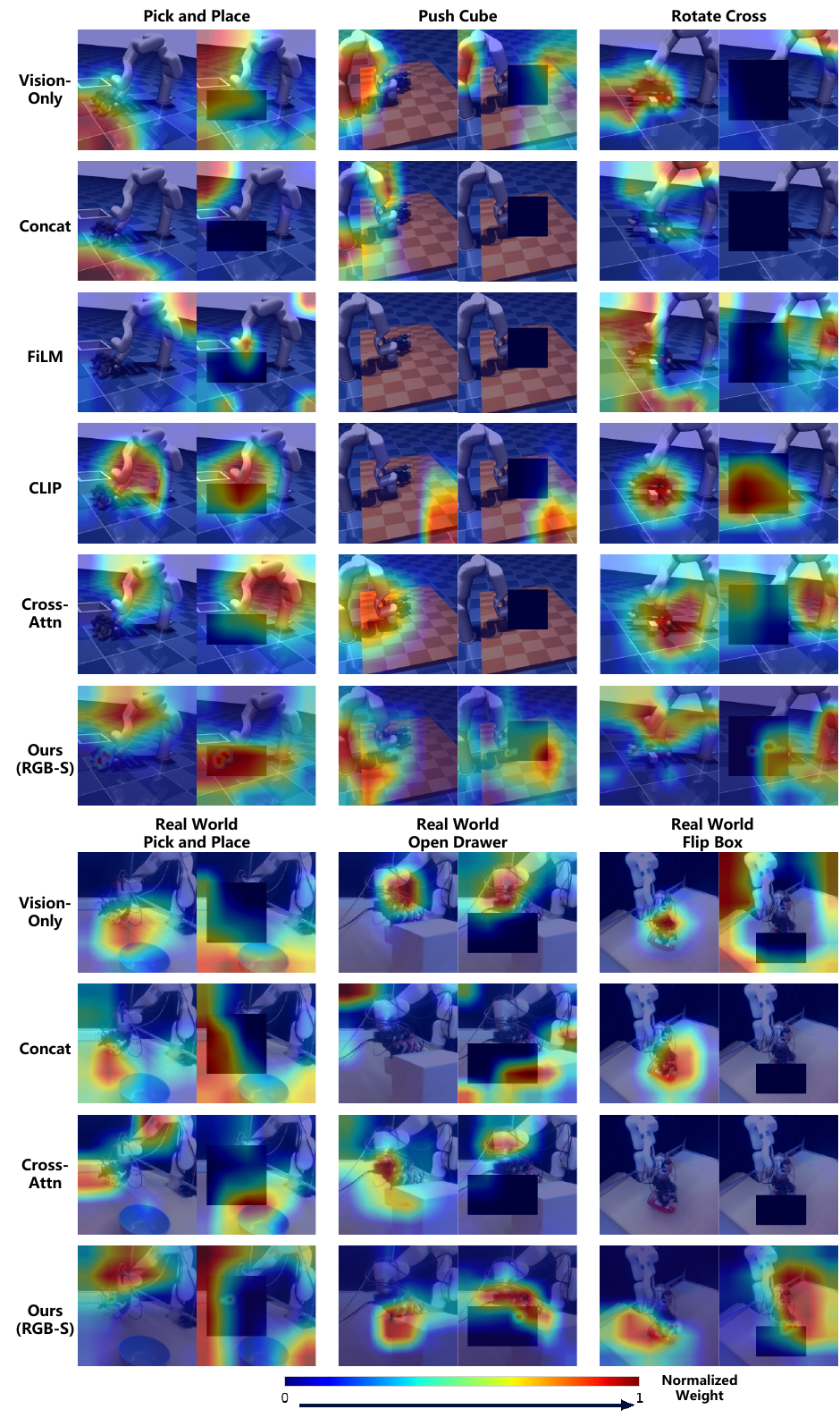}
    \caption{\textbf{Grad-CAM result of tasks in simulation and real-world.} }
    \label{fig:sim_gradcam}
\end{figure}
\clearpage

\section{Saliency Fusion Ablation Details}
\label{app:saliency_ablations}

We provide implementation details for the saliency-fusion ablation in Sec.~\ref{subsec:ablation_alignment}. 
All variants use the same Diffusion Policy backbone, training hyperparameters, RGB observations, proprioceptive states, and rendered saliency maps. 
The only difference is where the saliency stream is injected into the visual conditioning pipeline, which illustrated in Fig.~\ref{fig:ctrlnet_module}.

\paragraph{Early fusion without zero initialization.}
This variant uses exactly the same Tac2Pix architecture and training settings
as the full model, but initializes the added saliency-channel weights using
the standard initialization of the visual backbone rather than zeros.
Thus, the only changed factor is the initialization strategy.

\paragraph{Late fusion.}
The late-fusion variant treats the saliency map as an additional image-like modality. 
For each camera view, the RGB image and its corresponding saliency map are encoded by separate ResNet-18 encoders with the same spatial-softmax pooling interface. 
The saliency input is expanded to a 3-channel image to match the standard RGB encoder interface. 
The pooled RGB features, pooled saliency features, and other state inputs are concatenated after visual encoding to form the global condition for the diffusion U-Net. 
Thus, this variant increases the conditioning dimension but does not allow RGB and saliency features to interact before spatial pooling.

\paragraph{Intermediate fusion.}
The intermediate-fusion variant injects saliency features inside the ResNet visual encoder. 
The saliency map is first passed through a lightweight mask-projection branch, implemented as convolutional layers that map the single-channel saliency input to the channel dimension of an intermediate ResNet feature map. 
The projected saliency feature is then added to the corresponding RGB feature map through a zero-initialized projection layer before the remaining ResNet blocks are applied. 
This design follows a ControlNet-style residual conditioning formulation: the initial network behaves similarly to the RGB-only encoder, while the saliency residual is learned during fine-tuning~\citep{zhang2023addingconditionalcontroltexttoimage}. 
Compared with late fusion, this variant allows feature-level interaction between the two streams, but the saliency signal is introduced only after the earliest visual convolutions.

\begin{figure}[h]
    \centering
    \includegraphics[
        width=1\linewidth,
    ]{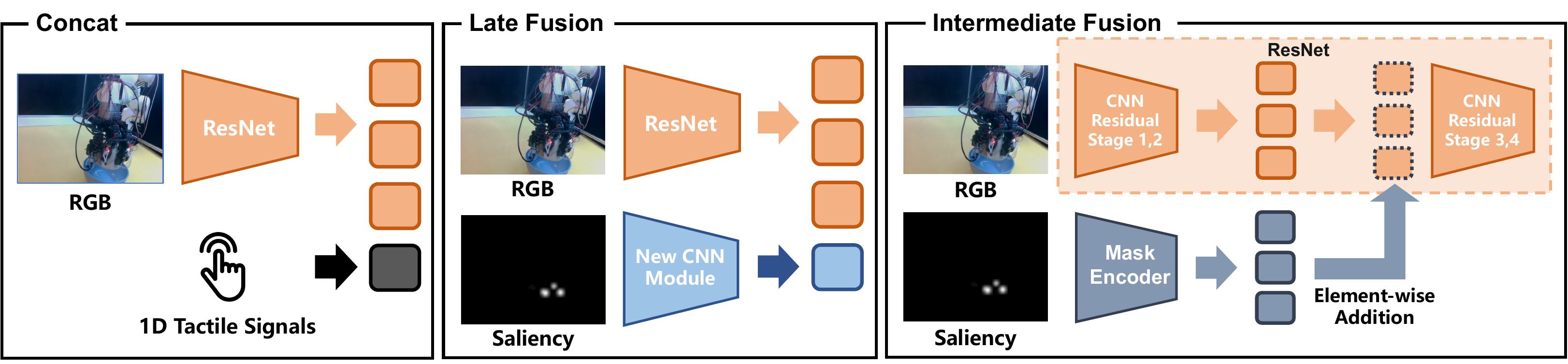}
    \caption{\textbf{Fusion architecture details.} (a) \texttt{concat}, where all features are concatenated to form the global conditioning vector. (b) Late-fusion and (c) intermediate-fusion variants.}
    \label{fig:ctrlnet_module}
\end{figure}

\section{Training Details of Fusing Mechanisms}
\label{app:fuse_mechms}

% FiLM modulation~\citep{perez2017filmvisualreasoninggeneral}, CLIP-style alignment~\citep{radford2021learningtransferablevisualmodels}, and cross-attention~\citep{jaegle2021perceiver}

\paragraph{Cross-attention fusion.}
\texttt{cross-attn} is implemented following~\citep{jaegle2021perceiver}. It represents multi-view images, tactile measurements, and the proprioceptive state as modality-specific tokens. All modality tokens are projected into a shared 128-dimensional hidden space and augmented with learnable view or modality embeddings. A set of four learnable latent queries attends to these tokens through 4-head cross-attention, followed by latent self-attention and a feed-forward block with a dropout rate of $0.1$. The resulting latent features are mean-pooled, concatenated with skip projections of the raw proprioceptive state and tactile vectors, and mapped to a 256-dimensional global conditioning vector for each observation step.

\paragraph{FiLM fusion.}

\texttt{FiLM} injects tactile information through two tactile-conditioned feature-wise modulation blocks, applied separately to the proprioceptive state feature and the concatenated multi-view visual feature. 
For a feature stream of dimension $d$, the corresponding FiLM block takes the tactile vector $f_t$ as input and predicts $2d$ modulation channels using a lightweight projection, $\mathrm{Linear}(\mathrm{Mish}(f_t))$. 
The output is split into a scale term $\gamma(f_t)\in\mathbb{R}^{d}$ and a bias term $\beta(f_t)\in\mathbb{R}^{d}$, and the feature is modulated as
$
\tilde{x}=\gamma(f_t)\odot x+\beta(f_t).
$
The final diffusion condition is formed by concatenating the tactile-modulated state feature, the tactile-modulated visual feature, and the tactile vector over the observation horizon.

\paragraph{CLIP-based fusion.}

\texttt{CLIP} is implemented following~\citep{radford2021learningtransferablevisualmodels} using a two-stage design. 
First, in the VT-CLIP pretraining stage, the vision encoder uses one ResNet-18 backbone per camera view, concatenates the 512-dimensional per-view features, and projects them to a 256-dimensional visual feature. 
The tactile encoder takes a flattened tactile-history window of length $5$ and maps it through a three-layer MLP with Mish activations to a 256-dimensional tactile feature. 
Both modalities are further projected to a 128-dimensional normalized contrastive space and trained with a symmetric CLIP loss. 
Second, during policy training, the pretrained VT-CLIP encoders are used as frozen feature extractors. 
The extracted features are concatenated with the proprioceptive state and flattened across the observation horizon to form the global conditioning vector.

\section{Training Details of Downstream Policies}
\label{app:downstream_policies}

\begin{table}[t]
\centering
\small
\setlength{\tabcolsep}{5.4pt}
\caption{Downstream policy and training hyperparameters used in our configurations.}
\label{tab:downstream_policy_hparams}
\begin{tabular}{cccccccccc}
\toprule
\makecell[c]{Policy} &
\makecell[c]{Domain} &
\makecell[c]{Obs.\\steps} &
\makecell[c]{Pred.\\horizon} &
\makecell[c]{Exec.\\steps} &
\makecell[c]{Batch\\size} &
\makecell[c]{Train\\steps} &
\makecell[c]{Optimizer} &
\makecell[c]{Learning\\rate} &
\makecell[c]{Weight\\decay} \\
\midrule
BC-MLP & Sim      & 1 & 1  & 1  & 64 & 120K & AdamW & $1\times10^{-4}$ & $1\times10^{-4}$ \\
ACT    & Sim      & 1 & 16 & 8  & 64 & 120K & AdamW & $1\times10^{-5}$ & $1\times10^{-4}$ \\
{DP} & {Sim/Real} & {4} & {24} & {8} & {64} & {120K} & {Adam} & {$1\times10^{-4}$} & {$1\times10^{-6}$} \\
{SmolVLA} & {Sim} & {1} & {50} & {10} & {8} & {120K} & {AdamW} & {$1\times10^{-4}$} & {$1\times10^{-10}$} \\
{3D-ViTac} & {Sim/Real} & {4} & {24} & {16} & {64} & {120K} & {AdamW} & {$1\times10^{-4}$} & {$1\times10^{-4}$} \\
{SaTA} & {Sim/Real} & {1} & {16} & {8} & {64} & {120K} & {AdamW} & {$1\times10^{-5}$} & {$1\times10^{-4}$} \\
\bottomrule
\end{tabular}
\end{table}

We evaluate four downstream policy backbones: BC-MLP~\citep{zhang2018deepimitationlearningcomplex}, ACT~\citep{zhao2023learningfinegrainedbimanualmanipulation}, DP~\citep{chi2024diffusionpolicyvisuomotorpolicy}, and SmolVLA~\citep{shukor2025smolvlavisionlanguageactionmodelaffordable} implemented based on Lerobot~\citep{cadene2026lerobot}. All policies use the same observation interface and action space if not mentioned in extra. Important training hyperparameters used in our configuration can be found in Table~\ref{tab:downstream_policy_hparams}. For controlled comparison, we use the same 120K optimization-update budget across policy architectures, which was sufficient for performance to largely stabilize in our experiments.

\paragraph{BC-MLP.}
\texttt{BC-MLP} is a single-step behavior cloning policy~\citep{zhang2018deepimitationlearningcomplex}. Each vector modality is projected to a 256-dimensional feature. The per-modality features are concatenated across the observation history and passed through an MLP with hidden dimensions $[512, 256]$, ReLU activation, dropout 0.1, and an MSE action regression loss.

\paragraph{ACT.}
\texttt{ACT} predicts an action chunk with a transformer encoder-decoder policy~\citep{zhao2023learningfinegrainedbimanualmanipulation}. The transformer uses a 512-dimensional hidden state, 8 attention heads, 4 encoder layers, and 1 decoder layer. Following the original ACT formulation, we enable the conditional VAE branch during training. The VAE encoder takes the ground-truth action chunk together with the proprioceptive state and encodes them into a 32-dimensional latent variable. The policy decoder is then conditioned on this latent variable, visual features, and the proprioceptive state to reconstruct the action chunk. The policy is trained with a L1 masked action reconstruction loss plus a KL regularization term. During inference, the VAE encoder is not used and the latent variable is set to the prior mean.

\paragraph{DP.}
\texttt{DP} predicts an action trajectory using a conditional 1D denoising U-Net~\citep{chi2024diffusionpolicyvisuomotorpolicy}. The diffusion U-Net uses channel dimensions $[512, 1024, 2048]$, a kernel size of $5$, group normalization with $8$ groups, a 128-dimensional diffusion-step embedding, and $100$ DDPM denoising steps. During deployment, we execute the first $8$ actions of each predicted action chunk and then replan with a new prediction.

\paragraph{SmolVLA.}
\texttt{SmolVLA}~\citep{shukor2025smolvlavisionlanguageactionmodelaffordable}
combines a pretrained SmolVLM2 backbone with a flow-matching action expert.
In our Tac2Pix implementation, the policy is conditioned on multi-view RGB
images, their corresponding tactile saliency maps, the proprioceptive
state, and a fixed task identifier.
Each saliency map is processed by a zero-initialized patch projection,
whose output is added to the corresponding RGB patch embeddings.
The pretrained vision-language backbone parameters remain frozen, while
the saliency projection, action expert, and input/output projection layers
are jointly optimized using the flow-matching objective.

\end{document}